%% file: modeling-guilt-emnlp2020.tex
\documentclass[11pt,a4paper]{article}
\usepackage[hyperref]{emnlp2020}
\usepackage{times}
\usepackage{enumitem}
\usepackage{amsmath}
\usepackage{latexsym}
\usepackage{multirow}

\usepackage{microtype}

\aclfinalcopy %

\usepackage{booktabs}
\usepackage{graphicx}
\usepackage[space]{grffile}  %

\newcommand{\Guilt}{SuspectGuilt}
\newcommand{\ReaderPerception}{\emph{Reader perception}}
\newcommand{\AuthorBelief}{\emph{Author belief}}
\newcommand{\cls}{\texttt{CLS}}
\newcommand{\word}[1]{\emph{#1}}

\definecolor{Orange}{RGB}{255,140,0}

\newcommand{\Secref}[1]{Section~\ref{#1}}
\newcommand{\secref}[1]{Section~\ref{#1}}

\newcommand{\Figref}[1]{Figure~\ref{#1}}
\newcommand{\figref}[1]{Figure~\ref{#1}}

\newcommand{\Tblref}[1]{Table~\ref{#1}}

\newcommand{\appref}[1]{Appendix~\ref{#1}}

\newcommand{\contribfootnote}[1]{%
  \begingroup
  \renewcommand{\thefootnote}{}\footnote{#1}%
  \addtocounter{footnote}{-1}%
  \endgroup
}

\title{Modeling Subjective Assessments of Guilt in Newspaper Crime Narratives}

\author{%
  Elisa Kreiss$^{1,*}$ \\\And
  Zijian Wang$^{2,*}$ \\[1ex]
  $^1$Department of Linguistics\quad$^2$Symbolic Systems Program \\
  Stanford University \\
  \texttt{\{ekreiss, zijwang, cgpotts\}@stanford.edu} \\\And
  Christopher Potts$^{1,2}$ \\
}

\date{}

\begin{document}
\maketitle
\begin{abstract}
\input{abstract.tex}\contribfootnote{$^{*}$Equal contribution.}
\end{abstract}

\section{Introduction}\label{sec:intro}
\input{introduction.tex}

\section{Related Work}\label{sec:related}
\input{related_work.tex}

\section{Data}\label{sec:data}
\input{data.tex}

\section{Models}\label{sec:model}
\input{model.tex}

\section{Experiments}\label{sec:experiment}
\input{experiment.tex}

\input{discussion.tex}

\section{Conclusion}\label{sec:conclusion}
\input{conclusion.tex}

\section{Acknowledgments}\label{sec:acknowledgments}
\input{acknowledgments.tex}

\bibliography{modeling-guilt-bib}
\bibliographystyle{acl_natbib}

\appendix

\section*{Appendices}
\input{appendices.tex}

\end{document}

%% file: abstract.tex
Crime reporting is a prevalent form of journalism with the power to shape public perceptions and social policies. How does the language of these reports act on readers? We seek to address this question with the \textbf{\Guilt\ Corpus} of annotated crime stories from English-language newspapers in the U.S. For \Guilt, annotators read short crime articles and provided text-level ratings concerning the guilt of the main suspect as well as span-level annotations indicating which parts of the story they felt most influenced their ratings. \Guilt\ thus provides a rich picture of how linguistic choices affect subjective guilt judgments. We use \Guilt\ to train and assess predictive models which validate the usefulness of the corpus, and show that these models benefit from genre pretraining and joint supervision from the text-level ratings and span-level annotations. Such models might be used as tools for understanding the societal effects of crime reporting.

%% file: introduction.tex
News outlets around the world routinely report on crimes and alleged crimes, ranging from petty misdemeanors to large-scale international criminal conspiracies. Each of these reports will frame events in ways that shape reader perceptions, and these perceptions will in turn shape public perception of how much crime there is, who is responsible for crime, and what policy decisions should be made to address crime. It is therefore important to understand how the language in these reports acts on readers, and there is clear value in developing NLP models that approximate these reader perceptions at a large scale, as a tool for estimating the aggregate effects of crime reporting on society.

\begin{figure}[tp]
  \centering

  \framebox{\includegraphics[width=0.95\linewidth]{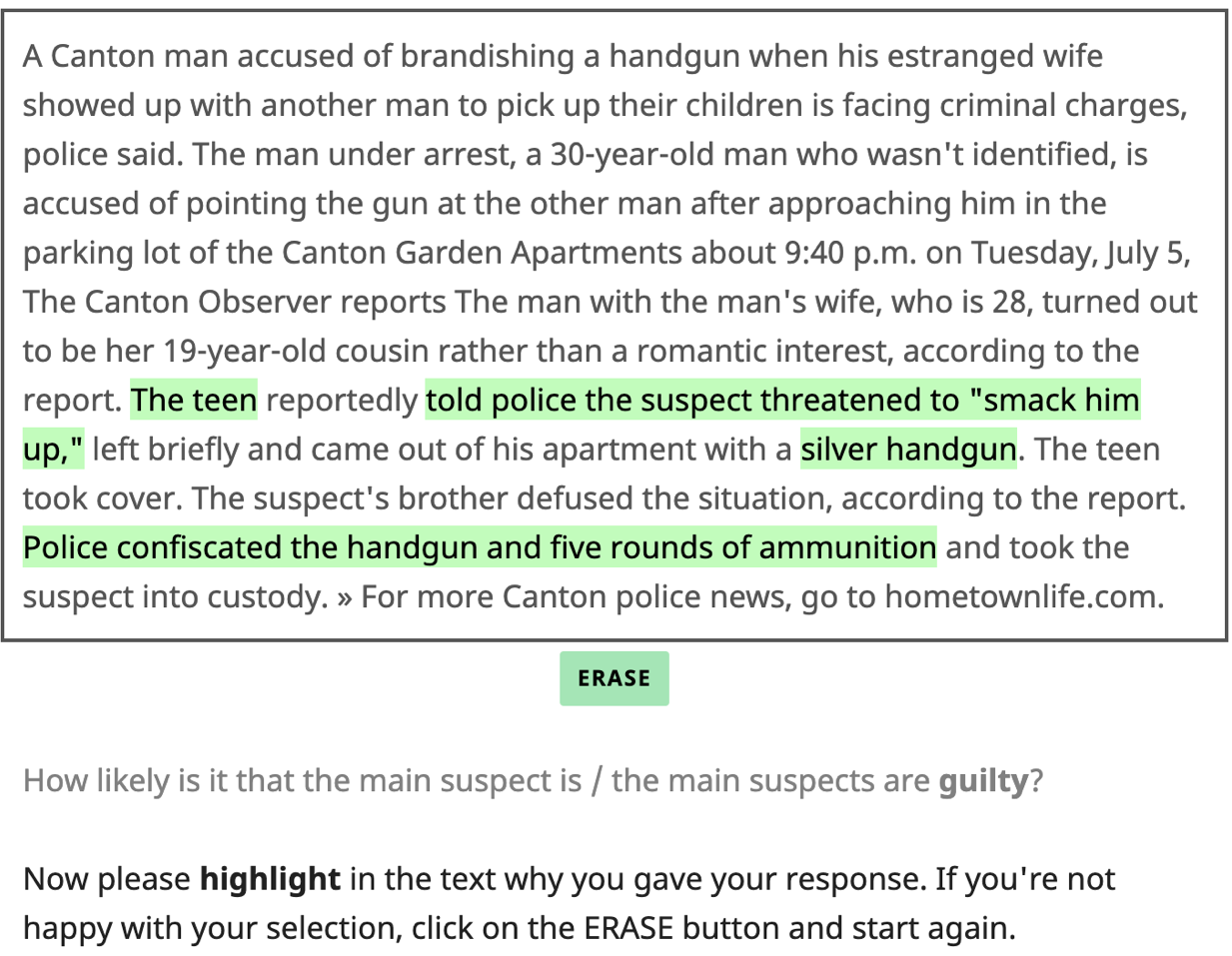}}
  \caption{The \Guilt\ corpus highlighting interface. After participants responded to a question about the guilt of the main suspect in the report, they completed this highlighting phase intended to provide insights into how they took themselves to be reasoning about the text. \Guilt\ contains 1.8K stories with at least 5 participants responding to each.}
  \label{fig:ann-highlighting}
\end{figure}

To begin to address these needs, we present the \textbf{\Guilt\ Corpus} of annotated crime stories from English-language newspapers in the U.S.\footnote{\url{https://github.com/zijwang/modeling_guilt}} Each story in the corpus is multiply-annotated with participants' assessments (on a continuous scale) of the guilt of the main suspect(s) and of the author's belief in the guilt of the suspect(s). In addition, for each of these guilt-rating questions, the participants highlighted the spans of text in the story that they felt contributed to their decision (\figref{fig:ann-highlighting}). These additional annotations provide a window into the language that participants took themselves to be attending to as part of their personal verdicts, and thus they are especially useful for understanding how authors' low-level linguistic choices feed into readers' overall judgments.

We also explore a range of methods for developing predictive models on the basis of \Guilt\ annotations which exemplify the usefulness of the resource. Our models are built on top of pretrained BERT parameters. In the simplest case, we learn to predict the author or subject guilt ratings without any other supervision. This basic model is improved if it is jointly trained on the guilt ratings and the span-level annotations that \Guilt\ provides, which helps to quantify the value of these low-level linguistic annotations. In addition, we explore unsupervised pretraining on a modestly-sized unlabeled corpus of crime stories, finding that it too increases the effectiveness of \Guilt\ models.

The span-level annotations offer new opportunities for analysis as well. Using the Integrated Gradients method of \citet{sundararajan2017axiomatic}, we identify the token-level features that our models rely on when trained without span-level supervision, and we compare this to the span-level annotations provided by \Guilt. Overall, the correspondence between the two is not high, which explains why the span-level objective helps our models and suggests that the document-level ratings alone might not suffice to yield models that attend to texts in the same ways that humans do.

%% file: related_work.tex
Our work draws on prior research into the relationship between language and assessments of guilt, as well as work seeking to jointly model text-level and token-level annotations using neural networks.

\subsection{Predicting Guilt}

The challenge of predicting guilt judgments from text sources has not yet received much attention. However, \citet{Fausey:Boroditsky:2010} show that using agentive language increases blame and financial liability judgments people make. Their results suggest that even subtle linguistic changes in crime reports will shape people's judgments of the events. More recent work has focused on predicting guilt verdicts from the Supreme Courts in the Philippines \citep{virtucio2018predicting} and Thailand \citep{kowsrihawat2018predicting} on the basis of presented facts and legal texts. \citeauthor{kowsrihawat2018predicting} employ a recurrent neural network with attention to make these predictions. These findings are for courtroom verdicts based on legal texts, and thus they are a useful complement to \Guilt, which provides subjective guilt judgments based on crime reporting.

\subsection{Veridicality Markers}\label{sec:markers}

We use the label `veridicality markers' to informally identify a large class of lexical items that includes hedges, evidentials, and other markers of (un)certainty. Analysis of the span-level annotations in \Guilt\ shows that veridicality markers play an out-sized role in shaping people's judgments of guilt. The annotations are dominated not only by conventionalized devices like \word{allegedly}, \word{suspect}, and \word{according to}, but also by more context-specific locutions like \word{police say} and \word{arrest}.

There is extensive prior literature on how veridicality markers affect the perceptions of the speaker and proposition \citep{Erickson-etal:1978, durik2008effects, bonnefon2006tactful, rubin:2007:ShortPapers, jensen2008scientific, ferson2015natural}. These studies suggest such markers affect people's judgments of credibility in differing ways. For example, an increase in the number of hedges decreases the credibility of witness reports \citep{Erickson-etal:1978} but increases the trustworthiness of journalists and scientists \citep{jensen2008scientific}. Additionally, the interpretation of hedges is context dependent \citep{bonnefon2006tactful,durik2008effects,ferson2015natural} and show high individual variation \citep{rubin:2007:ShortPapers,ferson2015natural}.

Similarly, attitude predications like \word{X reported S} can be used to reduce commitment, but they can also be used to provide evidence in favor of \word{S} (\citealt{Simons07,deMarneffe:Manning:Potts:2012,White:Rawlins:2018,White-etal:2018}). \citet{Stone:1994} and \citet{vonFintel:Gillies:2010} address similar uses of epistemic modal verbs. These findings show how complex these markers are pragmatically and highlight the value of usage-based studies of them.

\subsection{Span-Level Supervision}

BERT models \citep{Devlin:2018} define an output representation for every token-level input (see also \citealt{Vaswani:2017}). The parameters of these models can be fine-tuned in many ways \citep{lee2019mixout,mosbach2020stability}. Our models combine text-level prediction with sequence modeling; the supervision signals come from the guilt judgments and span highlighting in the \Guilt\ corpus. This basic model structure has been used in a wide variety of settings before. What is perhaps special about our use of it is that the two levels of annotation each provide evidence about the other; the highlighting can be seen as guiding the regression model to pay attention to certain words, and the regression label is likely to create helpful biases for particular token-level classifications. \citet{rei2019jointly} define models that similarly make use complementary tasks.
This is also conceptually very similar to the token-level supervision in the debiasing model of \citet{pryzant2020automatically}. However, while their token-level labels come from a fixed lexicon, ours were created in their linguistic context with a particular set of guilt-related issues in mind.

%% file: data.tex
The \Guilt\ corpus is a resource to investigate how the language of crime reports affects readers. This section describes the data collection and annotation process. We provide qualitative and quantitative analyses of \Guilt\ that exemplify its usefulness for psycholinguistic investigations and NLP applications.

\subsection{Data Collection} \label{sec:data-collection}

The \Guilt\ corpus is derived from a dataset of crime-related newspaper stories from regional, English-language newspapers in the U.S. We chose to focus on such stories because they are generally brief and self-contained.
By contrast, crime-related stories from major news outlets tend to involve public figures, political issues, and important global events, and readers' prior exposure to the issues might affect their judgments in unpredictable ways.

Inspired by \citet{davani2019reporting}, we collected our corpus from \href{http://www.patch.com}{Patch.com}. The Patch dataset contains independent, hyper-local news articles compiled from local news sites. We crawled all stories in the ``Crime \& Safety'' section for all news up through December 2019, yielding 474k news stories from 1,226 communities in the U.S. We then filtered this collection to just stories with (1) at most 300 words and (2) at least 4 of the following word-stems: \texttt{suspect*}, \texttt{alleg*}, \texttt{arrest*}, \texttt{crim*}, \texttt{accus*}\footnote{The word-stems were chosen to maximize the retrieval of news stories that report on criminal acts where a suspect has been identified but that still communicate uncertainty about the case.}. In addition, we filtered out stories that either have the same title, for which we only keep one copy, or are collections of multiple reports, e.g., records of incidents. As a post-processing step, we removed phone numbers and Patch.com advertisements. The final collection has 4.2K stories, of which we selected 1,957 for annotation.

\begin{figure}[tp]
  \centering
  \includegraphics[width=1\linewidth]{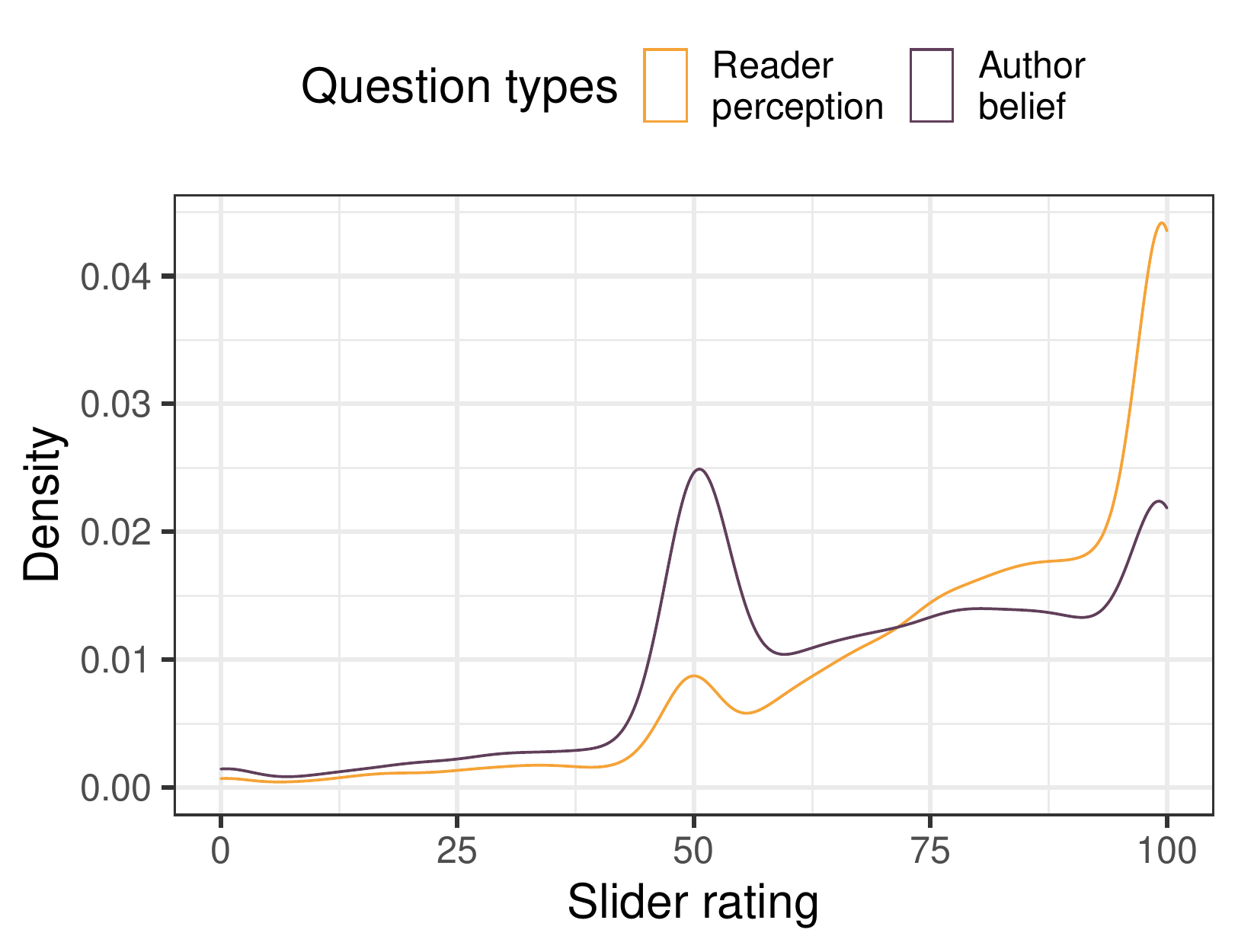}
  \caption{Slider rating density distribution for the \ReaderPerception\ and \AuthorBelief\ questions.}
  \label{fig:text-ann-dist}
\end{figure}

\subsection{Annotation Effort}\label{sec:data-ann}

For the annotation phase of \Guilt, participants were recruited on Amazon's Mechanical Turk and asked to read five stories and respond to three questions about them:
\begin{enumerate}\setlength{\itemsep}{0pt}
\item \ReaderPerception: ``How likely is it that the main suspect is / the main suspects are guilty?''
\item \AuthorBelief: ``How much does the author believe that the main suspect is / the main suspects are guilty?''
\item An attention check question, such as ``How likely is it that this story contains more than five words?''
\end{enumerate}
Responses were collected on a continuous slider, coded as ranging from 0 (very unlikely) to 1 (very likely). After submitting the slider response for each question, participants were asked to ``highlight in the text why [they] gave [their] response''. They additionally had the option to opt out of the slider response by indicating that the question didn't apply to the story. Stories with more than 30\% of ``Doesn't apply'' responses were excluded from the corpus, yielding 1,821 unique news reports.

\begin{figure*}[!ht]
  \centering
  \includegraphics[width=0.75\linewidth]{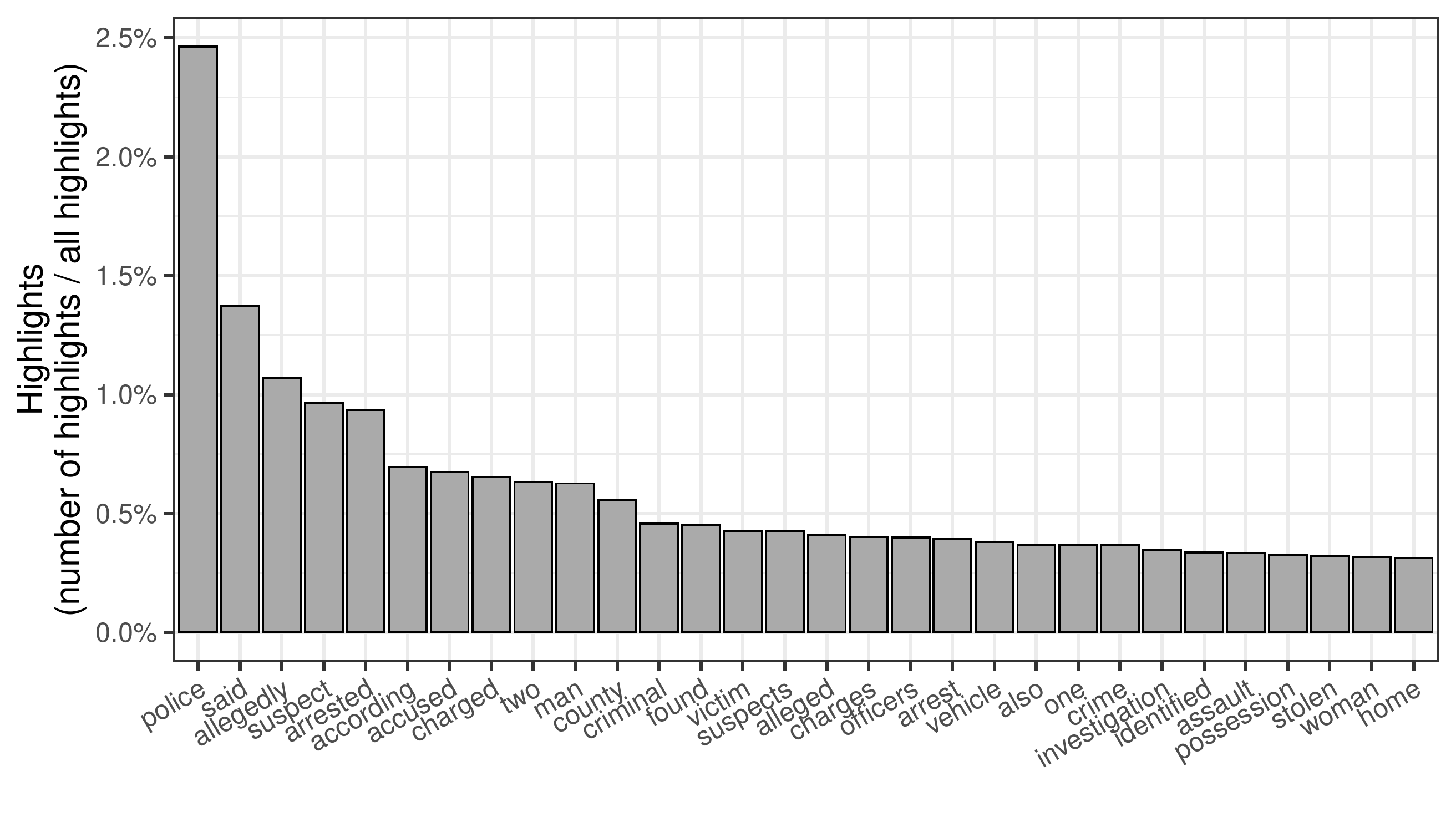}
  \vspace{-1mm}
  \caption{The 30 most highlighted words across questions.}
  \label{fig:highlighting-overall}
\end{figure*}

Guilt judgments are subjective and known to be highly variable (\secref{sec:markers}), and we expect the span-level highlighting to be even more variable. To accommodate this natural variation, we had multiple participants rate each story. Every story was annotated at least 5 times, and after excluding ``Doesn't apply'' responses, 99.2\% of the stories still have 5 annotations or more for the \ReaderPerception\ question and 86.7\% for the \AuthorBelief\ question. For our analyses and modeling in this paper, we generally average these annotations, but the corpus supports work at finer-grained levels. Our appendices include additional details, including screenshots of the annotation interface, exclusion criteria for participants, and aggregated participant demographics.

\begin{figure*}[!ht]
  \centering
  \includegraphics[width=0.75\linewidth]{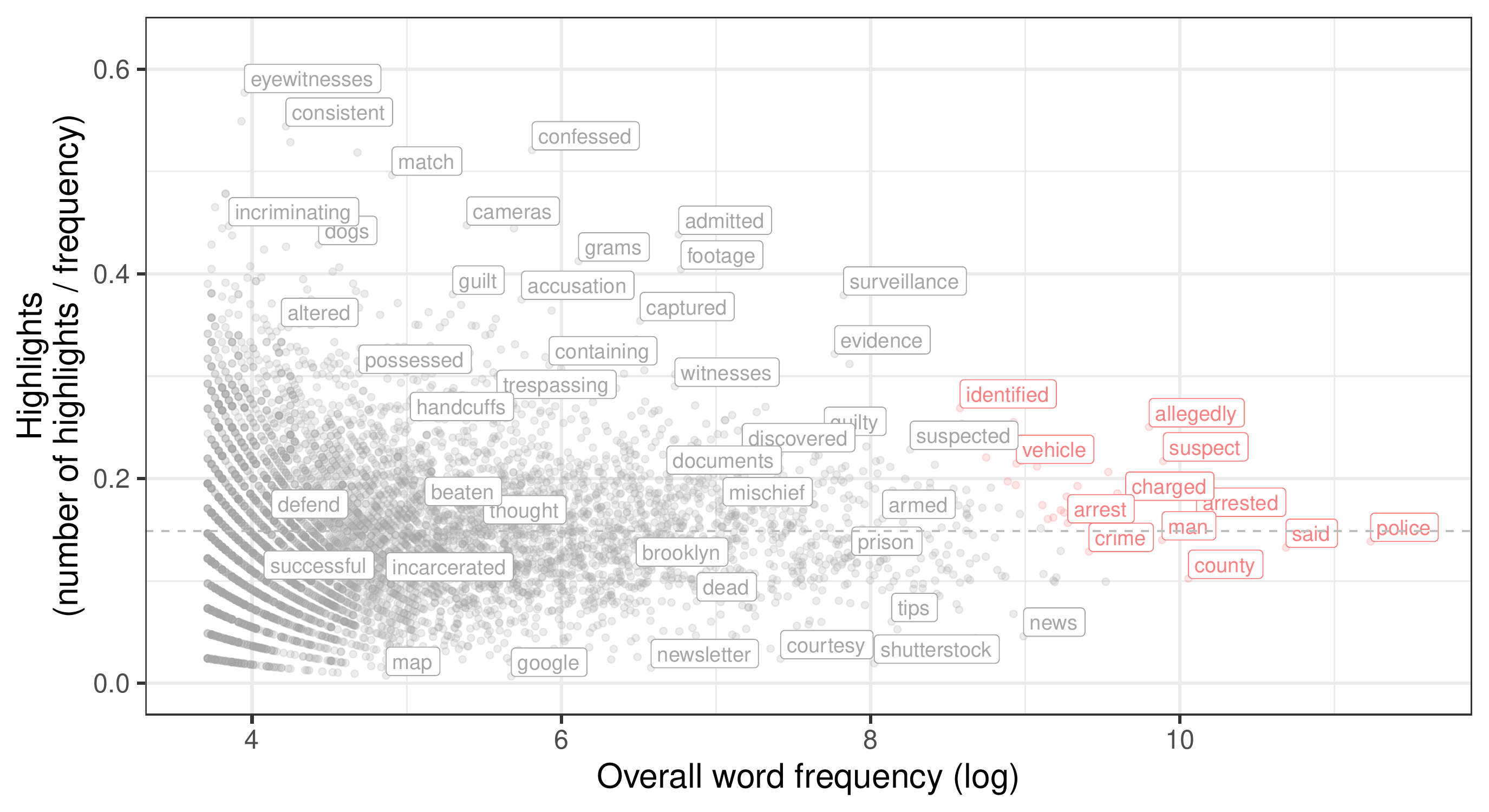}
   \vspace{-1mm}
  \caption{Proportion of token selections by frequency. Words that received the most highlights overall (see \Figref{fig:highlighting-overall}) are presented in red, other words in grey. By chance, words would be highlighted 14.88\% of the time, indicated by the dashed grey line. Words that are highlighted more often than predicted by chance are above this line, suggesting that they take on an important role in annotators' judgments.}
  \label{fig:highlighting-prop}
\end{figure*}

\subsection{Text-Level Annotations}\label{sec:data-guilt}

\Figref{fig:text-ann-dist} shows the distribution of responses for the \ReaderPerception\ and \AuthorBelief\ questions. Both distributions are skewed towards the middle and maximum portions of the slider scale. Relatively few participants chose ratings in the ``very unlikely'' range, which potentially reflects underlying biases about news reporting: readers expect suspects mentioned in these stories to be guilty. We also begin to see differences between the two questions. While \ReaderPerception\ ratings are rather skewed to the maximum portion of the scale, \AuthorBelief\ responses are concentrated around the center. This already suggests a disconnect between what readers believe about the suspect's guilt more generally and what readers believe about the author's beliefs. The cluster around the center also suggests that participants feel uncertainty, especially in the \AuthorBelief\ case. The clustering might also reflect a presumption that journalists will seek to appear unbiased.

We find high levels of interannotator agreement for both the \ReaderPerception\ and \AuthorBelief\ questions. The mean squared error (MSE) for each story is lower for the \ReaderPerception\ question (mean MSE = 0.0313) than \AuthorBelief\ (mean MSE = 0.0410). To provide some context for these numbers, we also calculated them after first shuffling all ratings. The MSE for this setting is 0.0443 for \AuthorBelief\ and 0.0353 for \ReaderPerception. Both are significantly higher than their non-shuffled counterparts according to a Welch Two Sample t-test ($p < 0.0001$).

\subsection{Span-Level Annotations}

When highlighting text spans, participants primarily marked passages shorter than 200 characters (approximately 33 words). \AuthorBelief\ highlights tended to be shorter than those for \ReaderPerception. Overall, highlights had a length between 1 and 1,717 characters (about 286 words). (A highlight here is defined as a consecutive mark without a non-highlighted character in between. If a participant highlighted two passages that are directly connected, they count as one highlighting.)

We would like to estimate agreement levels for span highlighting as well. Because our stories have varying numbers of annotations, we cannot calculate a Fleiss kappa value for this problem.
Krippendorff's alpha is a standard test that can accommodate this kind of variation, but its symmetric treatment of highlighting and non-highlighting is problematic since only 15\% of the tokens are highlighted.\footnote{Due to this inequality, the random baseline in Krippendorff's alpha (which is computed by shuffling each story's highlights) is disproportionately strong. The highlighting data still achieves a positive Krippendorff's alpha of 0.16 for \ReaderPerception\ and 0.08 for \AuthorBelief.} Nonetheless, to provide some insight into how alike our participants were in their highlighting behavior, we compared the percentage of annotators who highlighted each character with a random baseline. The random baseline highlights were created by randomly shuffling the underlying highlight distribution for each annotation.
We find that it was more likely that at least half of the annotators considered a token as important in the actual data as opposed to the random baseline (Welch Two Sample t-test: $p<0.0001$).

Token-level analysis of the highlighted spans reveals many connections with the markers of veridicality discussed in \secref{sec:markers}. \Figref{fig:highlighting-overall} shows the most highlighted words across the two guilt questions.\footnote{Punctuation and stopwords taken from the \textit{tm: Text mining package} in R were excluded for this analysis.} The list is dominated by conventionalized devices for signaling lack of commitment in newspaper reporting (e.g., forms of \word{allege}), devices for shifting attribution to others (e.g., \word{said}, \word{accused}), and genre-specific words that play into how we assess evidence in criminal contexts (e.g., \word{accused}, \word{charged}, \word{investigation}).

However, as we might expect, the number of times a word is highlighted highly correlates with its frequency ($r=0.97$). \Figref{fig:highlighting-prop} brings out this relationship. The x-axis is token frequency, and the y-axis gives the proportion of tokens for a word that were highlighted. (For example, if a word appeared 100 times and was highlighted 10 of those times, it would appear at 0.1 on the y-axis.) We excluded words with a frequency below 25, since these tend to get exaggerated proportions. The words from \figref{fig:highlighting-overall} are displayed in red and are highly frequent, and they are also the words with the highest highlighting proportion for their frequency, suggesting that these patterns are robust. Many of the other proportionally frequently highlighted words fall into the same categories as those in \figref{fig:highlighting-overall}: forms of \word{confess}, \word{eyewitnesses}, words picking out devices that provide evidence, and so forth. Words which were highlighted less than expected by chance (i.e., below the dashed grey line) rather reference meta-information of the news stories, such as \word{google}, \word{newsletter}, \word{shutterstock}, and \word{map}. In sum, the highlighting patterns seem aligned with the linguistic picture outlined in \secref{sec:markers}.

\begin{figure*}[!ht]
  \centering
  \includegraphics[width=0.75\linewidth]{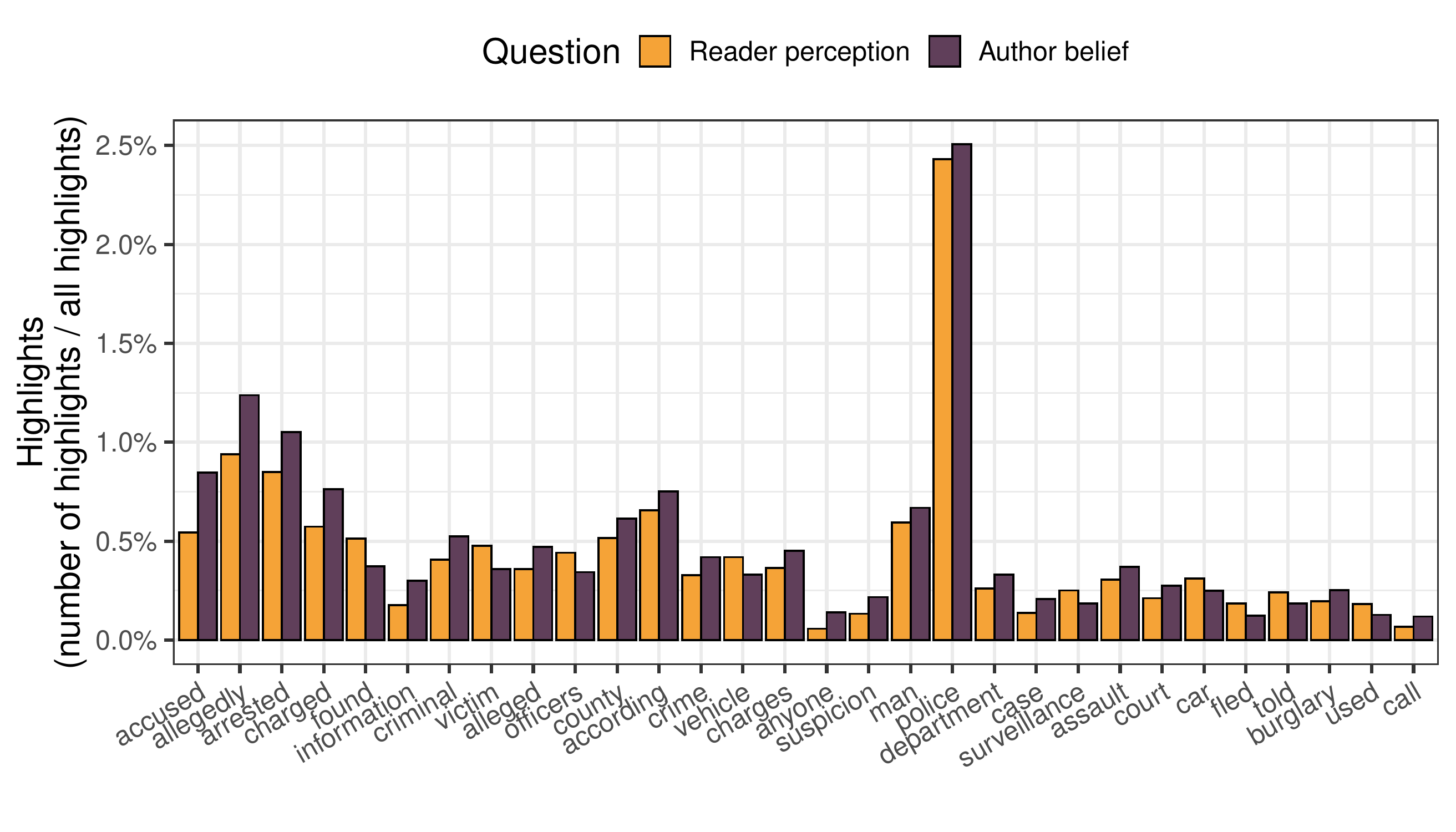}
  \caption{Words with the largest highlighting difference between the two guilt questions.}
  \vspace{-1mm}
  \label{fig:highlighting-conditions}
\end{figure*}

\Figref{fig:highlighting-conditions} seeks to add a further dimension to this analysis. Thus far, we have ignored the distinction between the two guilt-rating questions, \ReaderPerception\ and \AuthorBelief. The two questions are semantically quite different and might even come apart in some cases. For example, a reader might attend only to the evidence presented in a text and arrive at a high guilt-rating of their own, while ignoring clear indicators that the author wishes to remain non-committal about the origin or strength of that evidence. \citet{Kreiss:2019} found that hedges affect responses of \AuthorBelief\ but not \ReaderPerception, suggesting that the use of words like \word{allegedly} affects reader's perception about the author's beliefs but not their general guilt perception. This seems to be reflected in the selection data as well. In \figref{fig:highlighting-conditions}, we give the words with the largest differences between the two guilt questions. Conventionalized devices like these hedges, which signal lack of commitment in reporting, become even more prominent in the \AuthorBelief\ condition. This supports \citeauthor{Kreiss:2019}'s earlier findings of the relevance of these words for \AuthorBelief\ and not \ReaderPerception, and further suggests that readers appear to have some metalinguistic awareness for this difference.

%% file: model.tex
This section summarizes the family of models we consider in this work. All of them begin with BERT. We explore models with and without additional unsupervised pretraining on crime stories. We build regression models on top of these parameters using just the \cls\ token, which is the initial token in all BERT input sequences and is often taken to provide an aggregate sequence representation, as well as mean-pooling over all the final output states, and we additionally define extensions for predicting token-level highlighting.

\subsection{Guilt Ratings}\label{sec:model-BERT}

BERT \citep{Devlin:2018} is a Transformer-based architecture \citep{Vaswani:2017} that is usually trained jointly to do masked language modeling and next sentence prediction. The inputs are sequences of tokens $[x_{0}, \ldots, x_{n}]$, with $x_{0}$ designated as \cls\ and $x_{n}$ designated as \texttt{SEP}. BERT maps these inputs to a sequence of output representations $[h_{0}, \ldots, h_{n}]$.

Our two rating categories, \ReaderPerception\ and \AuthorBelief, define two separate tasks. We model them separately. For each, the core regression model is given by $hW_{r} + b_{r}$, where $W_{r}$ is a vector of weights, $b_{r}$ is a bias term, and $h$ is derived from the states  $[h_{0}, \ldots, h_{n}]$. In the \cls-based approach, $h = h_{0}$. In the mean-pooling approach, $h = \mathbf{mean}([h_{0}, \ldots, h_{n}])$.

The individual regression models are trained using a mean squared error (MSE) loss:
\begin{equation}
  \label{eq:regression}
  J_{r}(\theta_{r}) = \frac{1}{m} \sum_{i=1}^{m} \frac{1}{2} \|H_{\theta_{r}}(x_{i}) - y_{i}^{r}\|^{2}
\end{equation}
Here, $m$ is the number of examples, $\theta_{r}$ represents all the parameters of BERT plus our new task-specific  parameters $W_{r}$ and $b_{r}$, $y_{i}^{r}$ is the true label for example $x_{i}$, and $H_{\theta_{r}}(x_{i})$ is the prediction of the model for example $x_{i}$.

\subsection{Genre Pretraining}\label{sec:model-genre-pretraining}

BERT was trained on the BookCorpus \cite{zhu2015aligning} and Wikipedia. It often performs well on tasks involving very different data, but any domain shift has the potential to lower performance, and crime stories are a specialized genre. Previous work has shown that in-domain continued pretraining is often beneficial for end-task performance in such situations \citep[e.g.][]{han2019unsupervised, gururangan2020don}. We thus evaluate models with and without pretraining on unlabeled crime stories. For this, we use the unlabeled portion of the dataset described in \secref{sec:data-collection}.

\begin{figure*}[t!]
  \centering
  \includegraphics[width=0.97\linewidth]{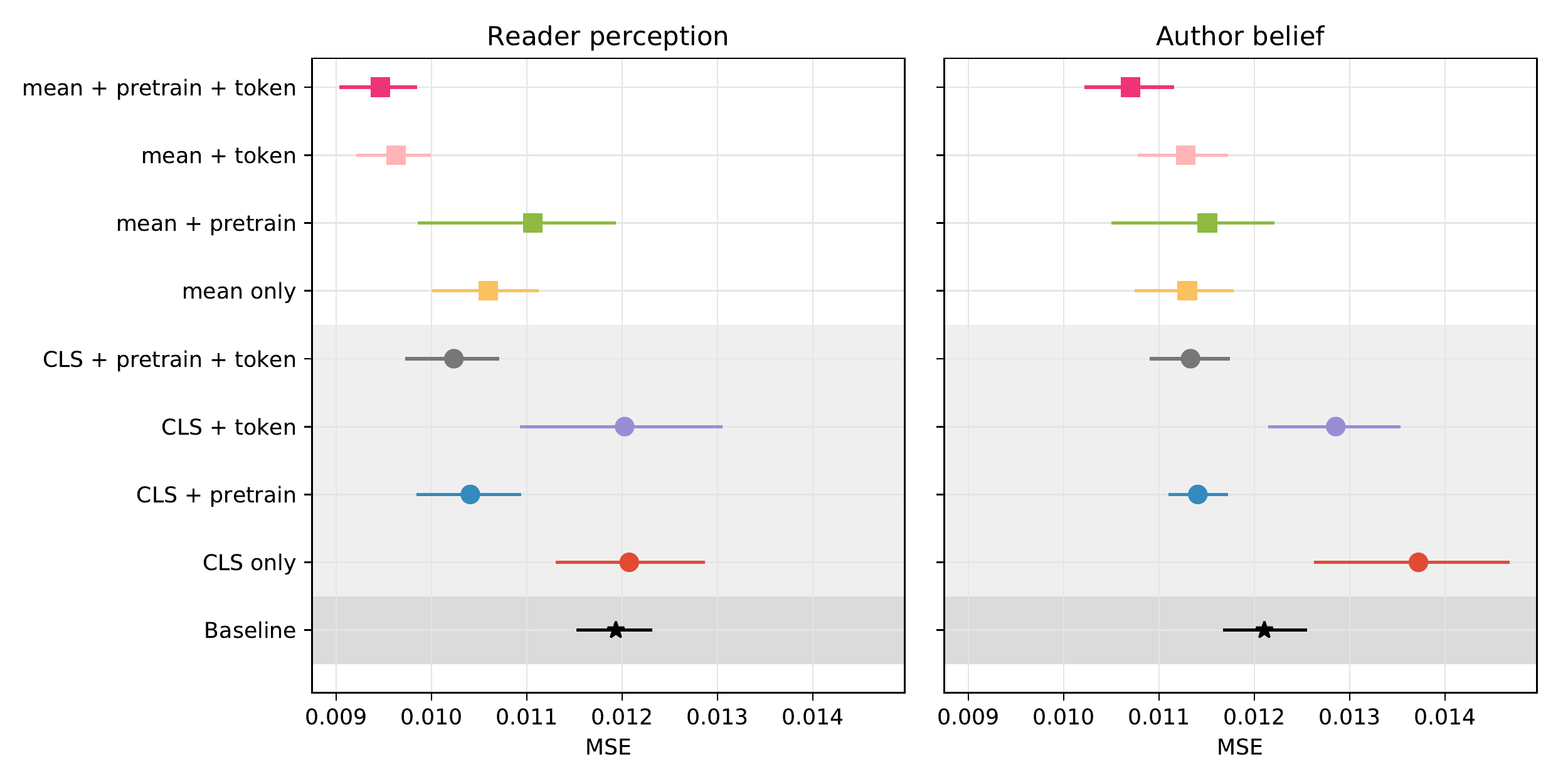}
  \caption{MSE (lower is better) for predicting guilt ratings for the \ReaderPerception\ and \AuthorBelief\ questions, with bootstrapped 95\% confidence intervals from 20 runs per model. `CLS' models use just the \cls\ token for the regression, whereas `mean' models average all the output heads (\secref{sec:model-BERT}). `pretrain' refers to genre pretraining (\secref{sec:model-genre-pretraining}), and `token' refers to token-level supervision from highlighting (\secref{sec:model-span-highlighting}).}
  \label{fig:results}
\end{figure*}

\subsection{Span Highlighting}\label{sec:model-span-highlighting}

We want to understand how authors' low-level linguistic choices affect readers' judgments of suspect guilt. To do this, we utilize the span-level annotation in \Guilt. Annotations are coded as 1 if the token was highlighted, and 0 otherwise. We merge the annotations of each news story to form a supplemental regression task, where the target value is the mean of the annotations. We use the output representation of each token from BERT and apply a linear regression similar to \eqref{eq:regression}:
\begin{equation}
\begin{split}
\label{eq:spans}
J_{t}(\theta_{t}) = \frac{1}{n}\frac{1}{m} \sum_{i=1}^{n} \frac{1}{2} \|H_{\theta_{t}}(x_{ij}) - y_{ij}^{t}\|^{2}
\end{split}
\end{equation}
Here, $m$ is the number of examples, $n$ is the number of tokens, and $x_{ij}$ and $y_{ij}$ stand for the $j$th token in example $i$, with corresponding token label  $y_{ij}^{t}$. $\theta_{t}$ denotes all the BERT parameters plus token-level regression parameters $W_{t}$ and $b_{t}$, and $H_{\theta_{t}}(x_{ij})$ is the prediction of the model for $x_{ij}$.

Our problem formulation might be taken to more naturally suggest a logistic regression. However, we opted for a linear regression objective instead, in the hopes that this would better capture not just the probability that a token is important, but also \emph{how important these tokens are}. The linear regression performed better in our evaluations, though the improvements over the logistic were modest.

\subsection{Joint Objective}\label{sec:model-objective}

The joint loss is a combination of the guilt-rating and span-highlighting objectives \eqref{eq:regression} and \eqref{eq:spans}:
\begin{equation}
\label{eq:3}
J = J_{r}(\theta_{r}) + \lambda J_{t}(\theta_{t})
\end{equation}
where $\lambda$ is a ratio of the losses that can be tuned.

%% file: experiment.tex
In this section, we report the evaluation procedure for the models described in \Secref{sec:model}. The results underline the usefulness of genre-pretraining and the rich annotations in the \Guilt\ corpus.

\subsection{Methods}\label{sec:experiment-predicting-guilt}

\begin{table}[tp]
  \centering
  \resizebox{0.75\linewidth}{!}{
    \begin{tabular}[t]{l c c}
      \toprule
      &  dev & test \\
      \midrule
      BERT-based & 2.224 & 2.223 \\
      Genre Pretrained & 0.884 & 0.887  \\
      \bottomrule
    \end{tabular}
  }
  \caption{Losses with and without genre pretraining.}
  \label{tab:genre-pretraining-ppl}
\end{table}

\begin{figure*}[!ht]
  \centering
  \includegraphics[width=0.76\linewidth]{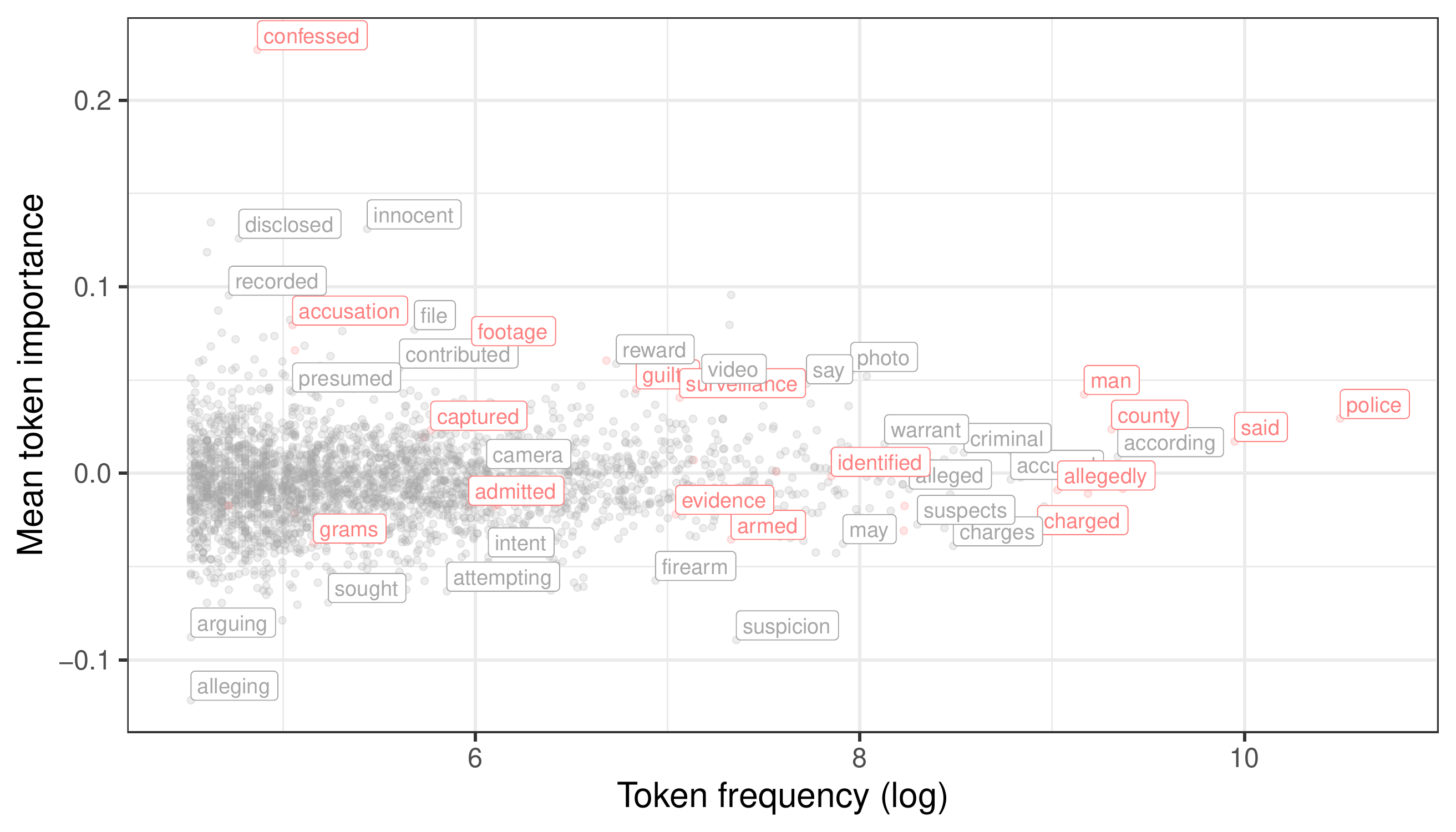}
  \caption{Mean model token importance by frequency. Words that received the most highlights overall (see \Figref{fig:highlighting-overall}) are presented in red, other words in grey. In contrast to the highlighting data in \Secref{sec:data-collection}, the token importance measure differentiates between tokens that increase (above 0 mean token importance) and decrease the predicted rating (below 0 mean token importance).}
  \label{fig:tokenimp}
\end{figure*}

We use the BERT-base uncased parameters for all of our experiments.

As discussed in \Secref{sec:model-genre-pretraining}, we performed pretraining with the $\approx$470K unlabeled articles from the dataset described in \secref{sec:data-collection}. We split the dataset into 80\% training, 10\% dev, and 10\% test sets. Additional details are  in \appref{sec:appendix-genre-pretraining}. \Tblref{tab:genre-pretraining-ppl} summarizes the quality of pretraining. The loss reduces up to 74\%, suggesting that genre pretraining could significantly improve in-domain performance. We evaluate the end-to-end performance of the genre pretraining next.

For the core guilt-rating prediction tasks, we split the \Guilt\ dataset into 85\% training and 15\% held-out test sets. We perform 5-fold cross validation and grid search on the training set. We then pick the best hyperparameters based on the best averaged loss of the 5-fold models, train our final model using the full training set for that fold, and report the final performance on the test set using the final model. We repeat the whole experiment with 20 different training-test splits to test the stability and significance of the performance. Additional details are given in \appref{sec:appendix-predicting-guilt}. We obtain a mean baseline by predicting everything as the mean values of the training set. We test the significance of \emph{whether A is better than B} using the Wilcoxon signed-rank test \cite{wilcoxon1992individual}.

\subsection{Results and Discussion}\label{sec:experiment-predicting-guilt-results}

Our results are summarized in \figref{fig:results}, which gives means and bootstrapped 95\% confidence intervals. (\Tblref{tab:results} in our appendix gives the precise numerical values with standard deviations, and expands on the statistical analyses.)

The results suggest that \AuthorBelief\ is a harder task than \ReaderPerception. This is aligned with the human results in \secref{sec:data-guilt}.

In general, the mean-pooling models are substantially better than the \cls-based ones. Indeed, we fail to find evidence that BERT with the \cls\ token improves performance over the baseline  ($p=0.440$ for \ReaderPerception; $p=0.996$ for \AuthorBelief). Furthermore, when using both genre pretraining and token supervision, mean pooling is also significantly better than using the \cls\ token ($p=0.001$ for \ReaderPerception; $p=0.022$ for \AuthorBelief).

Overall, a mean pooling model that makes use of genre pretraining as well as span-level supervision achieves the best performance. Span-level annotations are especially beneficial for the task of \AuthorBelief\ prediction, where this model significantly outperforms its closest competitors (e.g., when comparing against token supervision alone, $p=0.022$). We thus conclude that both token-level supervision and genre pretraining provide important information for \Guilt\ tasks.

%% file: discussion.tex
\section{Gradient-Based Token Importance}

Although our models predict human guilt judgments well, the performance metrics don't tell us \emph{how} they make predictions. Do they use information similarly to what we see in the human highlighting? Recent gradient-based methods for assessing feature importance in models like BERT \citep{sundararajan2017axiomatic,shrikumar2017learning} can help us answer this question.

\Figref{fig:tokenimp} presents one analysis of this form. We ran the Integrated Gradients method of \citeauthor{sundararajan2017axiomatic}, as implemented in the PyTorch Captum library, on models which received genre pretraining but no highlighting supervision. The figure includes test-set runs averaged across 20 models with different random train--test splits. A positive score means that the token increases the predicted rating; a negative score corresponds to a decrease.

Like our highlighting data, the neural network's importance scores show the highest variance for words with low frequency. Words that received higher highlighting proportions for their frequency primarily affect the model predictions positively. In addition, we find that words that are more likely than random to be highlighted (as described in \secref{sec:data}) are also significantly more likely to receive a higher token importance score in the model (Welch Two Sample t-test: $p<0.01$). Beyond this, however, there is little correlation between the absolute attribution score for each word and its highlighting proportion ($r=0.07$). While we can't rule out the possibility that this traces to the approximations introduced by Integrated Gradients, it seems likely that it helps explain why the span highlighting objective has a large impact on model predictions, as it is bringing in very different information than the model would otherwise attend to.

%% file: conclusion.tex
We introduced the \Guilt\ corpus, which provides a basis for a quantitative study of how readers arrive at judgments of \ReaderPerception\ and \AuthorBelief. We also showed that \Guilt\ can be used to train predictive models on top of BERT parameters, and that these models are improved by genre-specific pretraining and supervision derived from token-level highlighting.

Understanding how news reporting affects reader judgments is a difficult task. The span-level highlighting in \Guilt\ provides some insight into the factors at work here. We sought to match this with an introspective analysis of our predictive models using the gradient-based token importance method of \citet{sundararajan2017axiomatic}. This yielded a very different picture from what we see in \Guilt. Ultimately, this combination of annotations and model introspection might lead to new insights concerning how our models make decisions in this and other domains.

We also hope that this work paves the way to large-scale studies of how readers formulate judgments of guilt in crime reporting and encourages the development of systems that provide guidance on the presentation of these reports.

%% file: acknowledgments.tex
We thank the anonymous reviewers, Judith Degen, Daniel Lassiter, Michael Franke, and Sebastian Schuster for their generous comments and valuable suggestions on earlier versions of this work. Special thanks also to our Mechanical Turk workers for their essential contributions. This work is supported in part by a Google Faculty Research Award. Any remaining errors are our own.

%% file: appendices.tex
\section{Data}\label{sec:appendix-data}

\begin{figure*}[tp]
  \centering
  \includegraphics[width=1\linewidth]{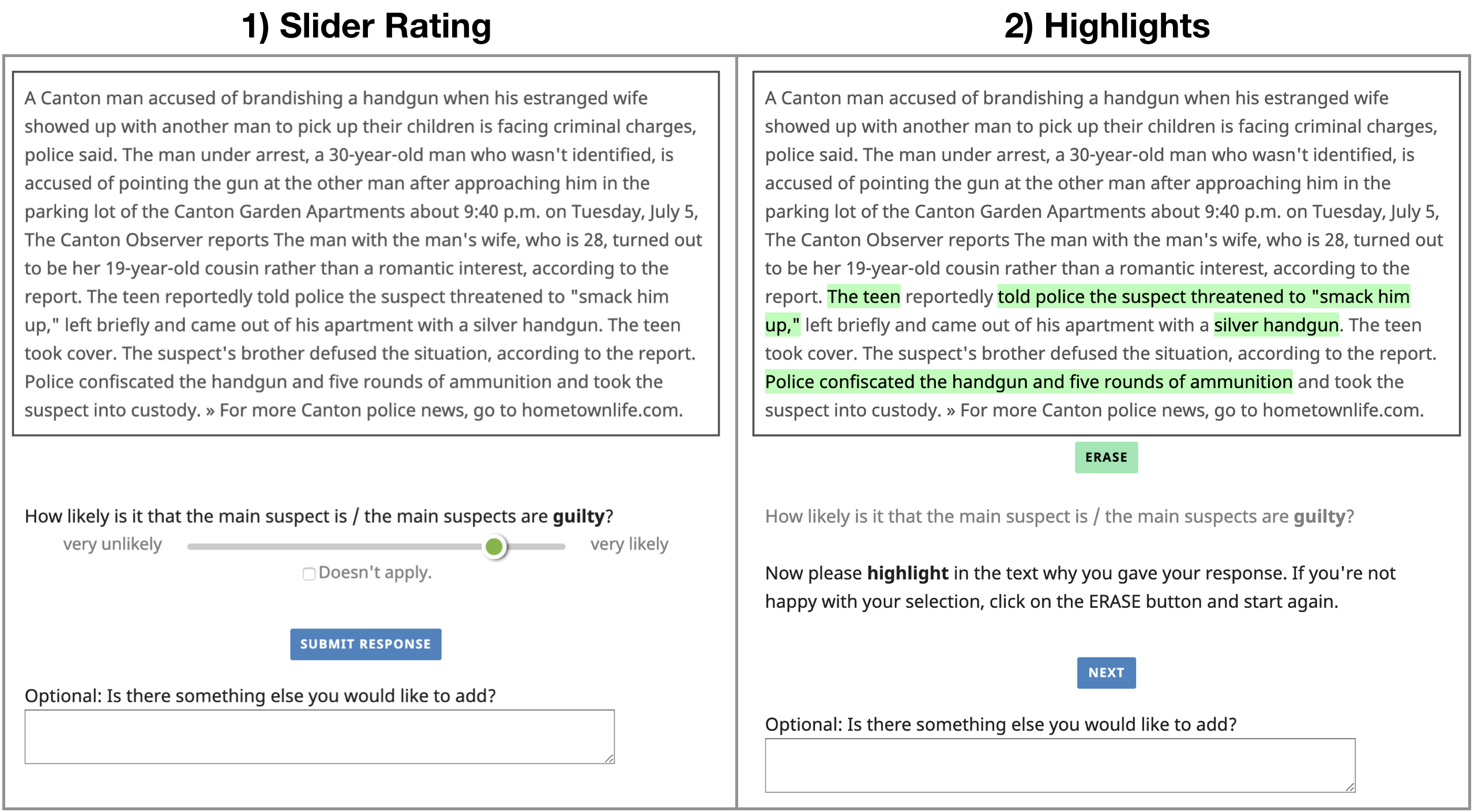}
  \caption{Participants rated a story on a continuous slider. After submitting, they highlighted the passages in the story that they considered to be most relevant for their assessment. At this point, they could not return to the previous screen to change the rating they gave.}
  \label{fig:design-app}
\end{figure*}

2,818 annotators contributed to 3,463 submissions on Amazon's Mechanical Turk. The approximate time for completion was 15 minutes, and each participant was paid \$2.50. We restricted participation to IP addresses within the US and an approval rate higher than 97\%. Participants were asked to read 5 stories and respond to three questions about them (as described in \Secref{sec:data-ann}). The full design of the trials is shown in \Figref{fig:design-app}.

We excluded participants who indicated that they did the study incorrectly or were confused (544), whose self-reported native language was not English (71), who spent less than 3.5 minutes on the task (53), and who gave more then 2 out of 5 erroneous responses in the control questions (359). A response is considered erroneous when a clearly true or false question incorrectly received a slider value below or above 50 (the center of the scale) respectively. Additionally, we excluded 120 annotations because annotators had seen this story in a previous submission. Overall, we excluded 1,035 submissions and 120 annotations (15,405 annotations out of 51,945, resulting in 36,420 annotations).

A majority of annotators (89\%) only participated once, which makes up 74\% of all annotations. Only 14 annotators participated more than three times (0.7\%).

The average age of annotators was 36 with a slightly higher proportion of male over female participants. The median time annotators spent on the study was 15.2 minutes, which is in-line with our original time estimates. Overall, annotators indicated that they enjoyed the study.

\begin{figure*}[tp]
  \centering
  \includegraphics[width=1\linewidth]{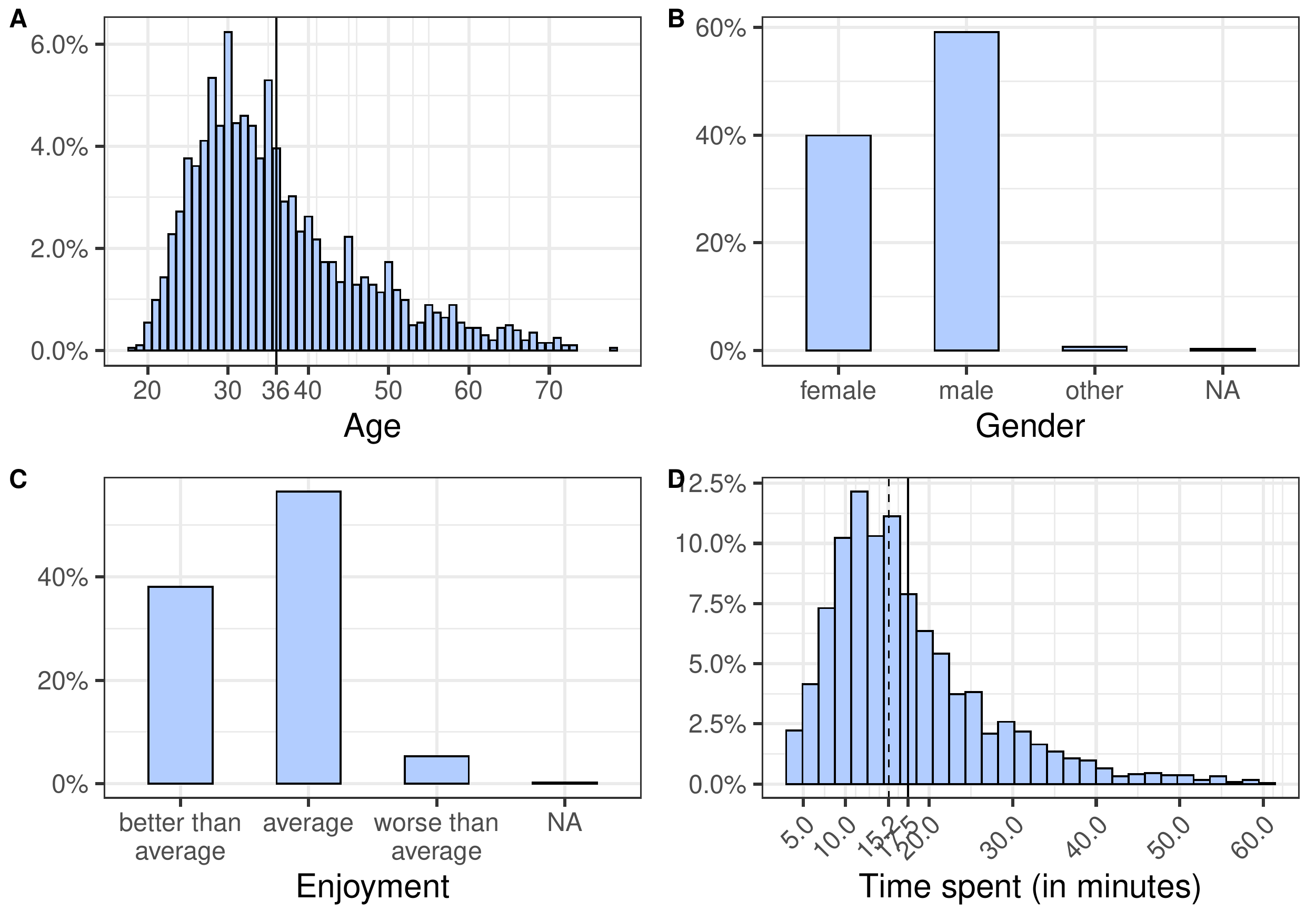}
  \caption{Participant demographics after exclusions.}
  \label{fig:part-demogr}
\end{figure*}

Annotators also had the option to indicate that the question cannot be applied to the news report. Overall, participants rarely used that option, but more so for the question about the \AuthorBelief\ (1.6\%) than the \ReaderPerception\ (10.5\%) question.
If several annotators agree that a question cannot be answered in the context of one particular story, it might be an indication that this story is not suitable for the corpus. We therefore decided to exclude stories where this box was selected more than 30\% of the time with that particular question. Further inspection showed that this mainly affected summary news articles which addressed multiple stories and suspects and therefore the questions could not be uniquely attributed to one specific case.

\section{Experiments} \label{sec:appendix-experiments}
\subsection{Genre Pretraining} \label{sec:appendix-genre-pretraining}
In this section, we describe the details of genre pretraining of BERT on our corpus. We set the maximum length to 400 tokens, with the tokens determined by the BERT tokenizer. This covers most of the instances in our corpus. We trained the model for 100K steps (roughly 30 epochs) using masked language modeling as described in \cite{Devlin:2018}, with a mask probability of 0.15, a batch size of 128, and a learning rate of $5 \cdot 10^{-5}$. All experiments throughout this paper are based on PyTorch \cite{paszke2019pytorch} and Huggingface's Transformers \cite{wolf2019HuggingFacesTS}.

\begin{table*}[tp]
  \centering
  \resizebox{0.7\linewidth}{!}{
    \begin{tabular}[t]{l cc}
      \toprule
      & \multicolumn{1}{c}{\ReaderPerception} &  \multicolumn{1}{c}{\AuthorBelief} \\
      & Mean \textsubscript{$\pm$std}  & Mean \textsubscript{$\pm$std}   \\
      \midrule
      Mean Baseline & $0.0119\,_{\pm 0.0009}$  & $0.0121 \,_{\pm 0.0010 }$ \\
      \midrule
      BERT (\texttt{CLS}) &  $0.0121 \,_{\pm 0.0018 }$ &   $0.0137 \,_{\pm 0.0025 }$  \\ %
      \hspace{1.5mm} + \textit{pretraining} & $0.0104 \,_{\pm 0.0013 }$  &   $0.0114 \,_{\pm 0.0007 }$   \\ %
      \hspace{1.5mm} + \textit{token supervision} & $0.0120  \,_{\pm 0.0024 }$  &  $0.0129 \,_{\pm 0.0015 }$ \\ %
      \hspace{1.5mm} + \textit{pretraining} + \textit{token supervision} &$0.0102 \,_{\pm 0.0011 }$ & $0.0113 \,_{\pm 0.0009 }$ \\  %
      \midrule
      BERT (\texttt{Mean}) & $0.0106 \,_{\pm 0.0013 }$  & $0.0113 \,_{\pm 0.0012 }$ \\
      \hspace{1.5mm} + \textit{pretraining} & $0.0111 \,_{\pm 0.0024 }$  &   $0.0115 \,_{\pm 0.0019 }$   \\
      \hspace{1.5mm} + \textit{token supervision} & $0.0096 \,_{\pm 0.0009 }$  &  $0.0113 \,_{\pm 0.0011 }$ \\
      \hspace{1.5mm} + \textit{pretraining} + \textit{token supervision} &\textbf{0.0095}$ \,_{\pm 0.0009 }$  & \textbf{0.0107}$ \,_{\pm 0.0011 }$ \\
      \bottomrule
    \end{tabular}
  }
  \caption{MSE for predicting guilt ratings for the \ReaderPerception\ and \AuthorBelief\ questions. The models themselves are defined in \secref{sec:model}. We report the mean and standard derivation values from 20 different runs. Bold denotes the best performance.}
  \label{tab:results}
\end{table*}

\subsection{Predicting Guilt} \label{sec:appendix-predicting-guilt}
In this section, we describe the hyperparameters used in our experiment.

For the basic models where there is no token supervision, we use the following hyperparameters
\begin{itemize}[noitemsep,topsep=2pt]
	\item Number of epochs: 5
	\item Warmup ratio: 10\%
	\item Learning rate: 3E$-$5, 5E$-$5
	\item Random Seed: 0, 1
	\item Batch size: 16
	\item Checkpoints: 100 steps per checkpoint
\end{itemize}
We also experimented with different number of epochs, batch sizes, and oversampling tail cases with different ratios in an initial small-scale study. We found that the current set of hyperparameters performs well in general. As adding more hyperparameter options is computationally intensive, we decided to use this set for our full-scale experiments.

When training the final model, we use the checkpoint whose corresponding steps are closet to 1.25 times the average number of steps of best performing checkpoints in the 5-fold cross validation.

For the models with token supervision, we use the same set of hyperparameters of no token supervision models except we only use one seed and add a hyperparameter of the loss ratio $\lambda$, with options of $[1,2]$.

\subsection{Numerical Results}

\Tblref{tab:results} gives the corresponding numerical values for \figref{fig:results}. Whereas \figref{fig:results}  gives bootstrapped confidence intervals, here we given standard deviations to quantify the amount of variation seen across runs. Below are some additional details on these comparisons (`AB' = \AuthorBelief; `RP' = \ReaderPerception. Our statistical test here is the Wilcoxon signed-rank test.)
\begin{enumerate}
\item BERT with the \cls\ token does not improve performance compared to a simple mean baseline  ($p=0.449$ for RP and $p=0.998$ for AB), while BERT with mean-pooling achieves better performance compared to the mean baseline ($p<0.001$ for RP and $p=0.004$ for AB).

\item The differences between using mean pooling and the \cls\ token are significant ($p=0.003$ for RP and $p<0.001$ for AB).

\item When using both the genre pretraining and the token supervision, mean pooling is significantly better than using the \cls\ token ($p=0.001$ for RP and $p=0.022$ for AB).

\item Overall, a mean pooling model that makes use of genre pretraining as well as span-level supervision achieves the best performance, significantly outperforming other models ($p<0.001$ for RP and  $p=0.027$ for AB when comparing with the mean baseline; $p=0.001$ for RP and  $p=0.020$ for AB with genre pretraining; and $p=0.131$ for RP and $p=0.022$ for AB with joint supervision).

\item Neither mean pooling models with genre pretraining ($p=0.649$ for RP and $p=0.464$ for AB) nor span-level supervision ($p=0.001$ for RP and $p=0.215$ for AB) alone can improve performance substantially in comparison to the mean baseline (only joint supervision for RP is significant).
\end{enumerate}